\documentclass{article}

\usepackage{microtype}
\usepackage{graphicx, stfloats}
\usepackage{subfigure}
\usepackage{booktabs} 

\usepackage{hyperref}

\usepackage[accepted]{icml2023}

\usepackage{amsmath}
\usepackage{amssymb}
\usepackage{mathtools}
\usepackage{amsthm}

\usepackage[capitalize,noabbrev]{cleveref}

\theoremstyle{plain}

\theoremstyle{definition}

\theoremstyle{remark}

\usepackage[textsize=tiny]{todonotes}

\icmltitlerunning{PQV-Mobile: A Combined Pruning and Quantization Toolkit to Optimize Vision Transformers for Mobile Applications}

\begin{document}

\twocolumn[
\icmltitle{PQV-Mobile: A Combined Pruning and Quantization Toolkit to \\ Optimize Vision Transformers for Mobile Applications}



\icmlsetsymbol{equal}{*}

\begin{icmlauthorlist}
\icmlauthor{Kshitij Bhardwaj}{yyy}
\end{icmlauthorlist}

\icmlaffiliation{yyy}{Lawrence Livermore National Lab}

\icmlcorrespondingauthor{Kshitij Bhardwaj}{Bhardwaj2@llnl.gov}

\icmlkeywords{Machine Learning, ICML}

\vskip 0.3in
]



\printAffiliationsAndNotice{}  

\begin{abstract}

While Vision Transformers (ViTs) are extremely effective at computer vision tasks and are replacing convolutional neural networks as the new state-of-the-art, they are complex and memory-intensive models. In order to effectively run these models on resource-constrained mobile/edge systems, there is a need to not only compress these models but also to optimize them and convert them into deployment-friendly formats.
To this end, this paper presents a combined pruning and quantization tool, called PQV-Mobile, to optimize vision transformers for mobile applications. The tool is able to support different types of structured pruning based on magnitude importance, Taylor importance, and Hessian importance. It also supports quantization from FP32 to FP16 and int8, targeting different mobile hardware backends. We demonstrate the capabilities of our tool and show important latency-memory-accuracy trade-offs for different amounts of pruning and int8 quantization with Facebook Data Efficient Image Transformer (DeiT) models. Our results show that even pruning a DeiT model by 9.375\% and quantizing it to int8 from FP32 followed by optimizing for mobile applications, we find a latency reduction by $7.18\times$ with a small accuracy loss of 2.24\%. The tool is open source.
\footnote{ \url{https://github.com/kshitij11/PQV-Mobile}}
\end{abstract}

\section{Introduction}
\label{sec:intro}

Vision Transformers (ViTs)~\cite{kolesnikov2010image} have recently emerged as a competitive alternative to Convolutional Neural Networks (CNNs) that are currently state-of-the-art in different image recognition computer vision tasks. ViTs have shown to outperform the CNNs by almost $4\times$ in terms of computational efficiency and accuracy~\cite{viso}. Additionally, ViTs have been shown to be more robust than CNNs and can be easily trained on smaller datasets. 

While ViTs are extremely effective at computer vision tasks, they are complex and memory-intensive models. For example, Facebook's Data Efficient Image Transformers (DeiT) take 331MB memory and are therefore not suitable for resource-constrained edge systems such as for mobile applications. Previous research has focused on pruning~\cite{fang2023depgraph} and quantizing~\cite{li2023vit} ViTs, but mostly separately and they do not target mobile applications where deployment of such models is challenging and requires converting these models to mobile hardware friendly lightweight and optimized formats.

This paper presents a combined pruning and quantization tool, called PQV-Mobile, to optimize vision transformers for mobile applications. The tool is able to support different types of structured pruning based on magnitude importance, Taylor importance, and Hessian importance. It also supports quantization from FP32 to FP16 and int8, tailored towards several hardware backends, such as x86, FBGEMM (Facebook General Matrix Multiplication~\cite{fbgemm}), QNNPACK (Quantized Neural Network Package~\cite{qnnpack}), and ONEDNN~\cite{onednn}. The pruned and quantized models are optimized for mobile applications and converted to mobile-friendly lightweight formats. We demonstrate the capabilities of our tool and show important latency-memory-accuracy trade-offs for different amounts of pruning, int8 quantization, and hardware backends with two types of Facebook DeiT models.

Our results show that even pruning a DeiT model by 9.375\% and quantizing it to int8 from FP32, we find a latency reduction by $7.18\times$ with a small accuracy loss of 2.24\%. All of our compared models are optimized for mobile applications and converted into deployment friendly lightweight formats.

\section{PQV-Mobile Tool}
\label{sec:tool}

\begin{figure}[t]
    \centering
    \includegraphics[width=0.7\columnwidth]{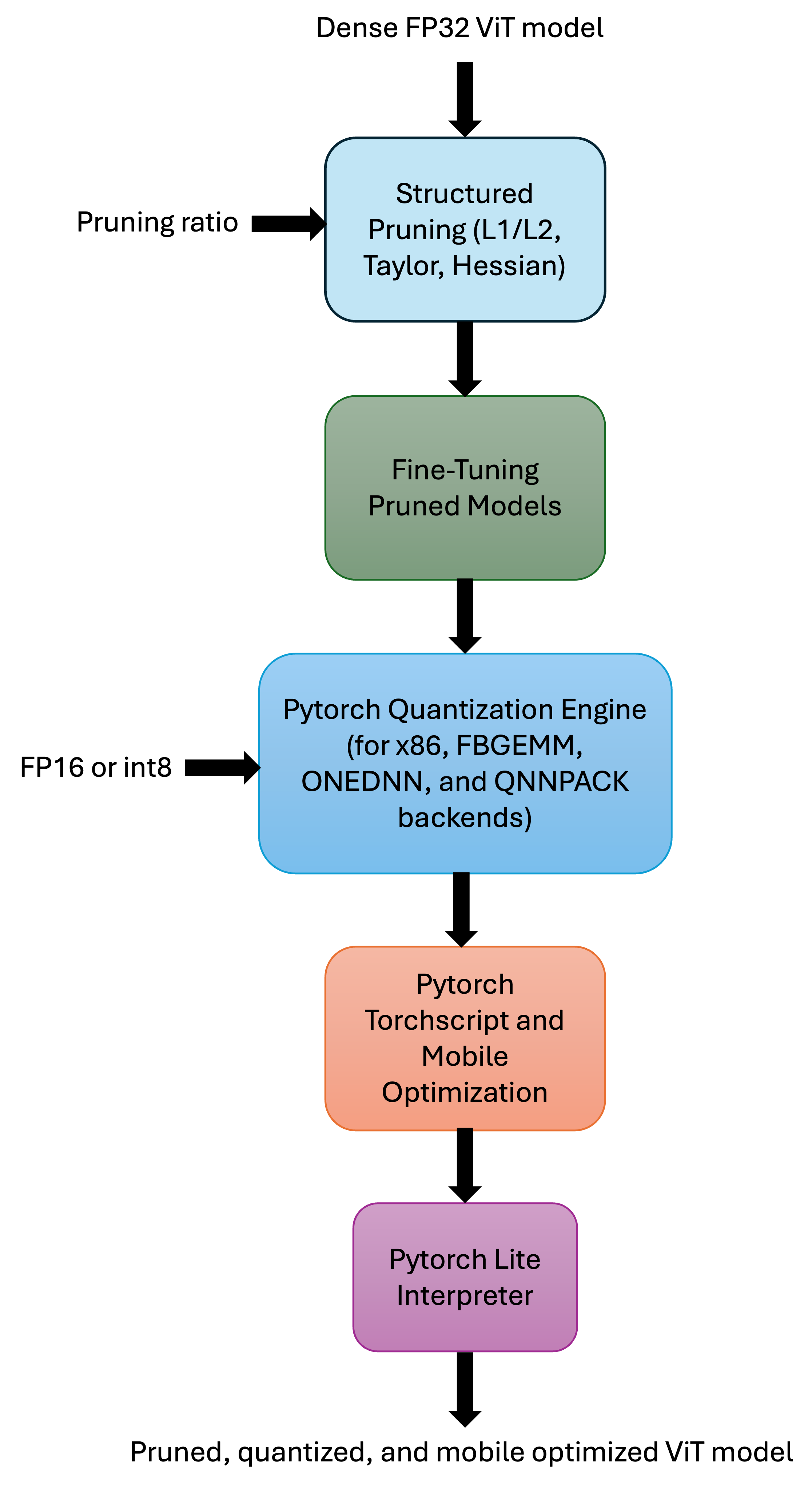}
    \caption{PQV-Mobile tool flow\vspace{-0.2in}}
    \label{fig:flow}
\end{figure}

Figure~\ref{fig:flow} shows our PQV-Mobile Tool flow. It supports different kinds of post-training pruning strategies such as L1, Taylor, etc. We found that the pruned model shows a major accuracy drop and therefore needs to be finetuned. The pruned and finetuned model is then input to a quantization engine, which can be tailored towards various hardware backends, and optimized for mobile applications. Our results showed that there is a small accuracy drop after quantization and therefore we did not perform any further finetuning of the quantized model. Currently, our tool can handle any of the HuggingFace ViTs from the TIMM library (Pytorch Image Models)~\cite{timm}. This section describes these steps in more details.

\subsection{Pruning method}
\label{subsec:pruning}

PQV-Mobile supports several pruning strategies, corresponding to structured pruning. In structured pruning, a block is removed, which can be a neuron in a fully-connected layer, a channel of filter in a convolutional layer, or a self-attention head in a Transformer. An alternative approach is unstructured pruning (also called magnitude pruning) where some of the parameters or weights with smaller values are converted to zeroes. PQV-Mobile targets structured pruning methods as they do not rely on specific AI accelerators
or software to reduce memory consumption and computational costs, thereby finding a wider domain of applications in practice~\cite{fang2023depgraph}. 

In structural pruning, a `Group' is defined as the minimal unit that can be removed. Many of these groups consist of multiple layers which can be interdependent and need to be pruned together so as to maintain the integrity of the resulting pruned networks. We follow the approach of~\cite{fang2023depgraph} that uses a dependency graph to model these dependencies and find the right groupings for parameter pruning. Similar to~\cite{fang2023depgraph}, PQV-Mobile accepts a group (i.e., an Attention block of a ViT with Linear layers) as inputs, and returns a 1-D tensor with the same length as the number of channels. All groups must be pruned simultaneously and thus their importance should be accumulated across channel groups. PQV-Mobile supports the following groupings:

{\bf Magnitude importance based grouping.} In this case, L1- or L2-norm regularization term is applied to the loss function which penalizes non-zero parameters. If the value of a connection is less than a threshold, the connection is dropped~\cite{anwar2017structured}.  

{\bf Taylor importance based grouping.} The importance is calculated as the squared change in loss induced by removing a specific filter from the network. This importance is approximated with a Taylor expansion which allows for faster computation from parameter gradients, even for larger networks~\cite{molchanov2019importance}.

{\bf Hessian importance based grouping.} In this method the importance is computed using a fast second-order metric to find insensitive parameters in a model. In particular, the average
Hessian trace is used to weight the magnitude of the parameters; parameters with large second-order sensitivity remain unpruned, and those with relatively small sensitivity are pruned~\cite{yu2022hessian}.

\subsection{Quantization method} PQV-Mobile currently supports post-training quantization of both weights and activations from FP32 to FP16 and int8. We plan to extend this to int4 as future work. The following steps are performed for quantization using quantization libraries of Pytorch:
\begin{itemize}
    \item Quantize models for a specific backend: We first create a quantization engine based on a backend. The supported backends are: x86, FBGEMM (Facebook General Matrix Multiplication~\cite{fbgemm}), QNNPACK (Quantized Neural Network Package~\cite{qnnpack}), and ONEDNN~\cite{onednn}. The generated engine is then used to quantize the model using either static or dynamic quantization~\cite{quant}.
    \item Convert Pytorch models to Torchscript format: Python models are inefficient to run during deployment. Therefore, we export the Pytorch models to production environments through Torchscript~\cite{jit}, which is an easy way to create serializable and optimizable models.
    \item Use Pytorch's mobile optimizer to optimize the quantized and scripted model~\cite{mobile} for mobile applications.
    \item Use Pytorch's Lite Interpreter to create a deployable and light version of the mobile optimized model~\cite{lite}.
\end{itemize}

\section{Experimental Results}
\label{sec:results}

\begin{figure}[t]
    \centering
    \includegraphics[width=0.9\columnwidth,trim=4 4 4 4,clip]{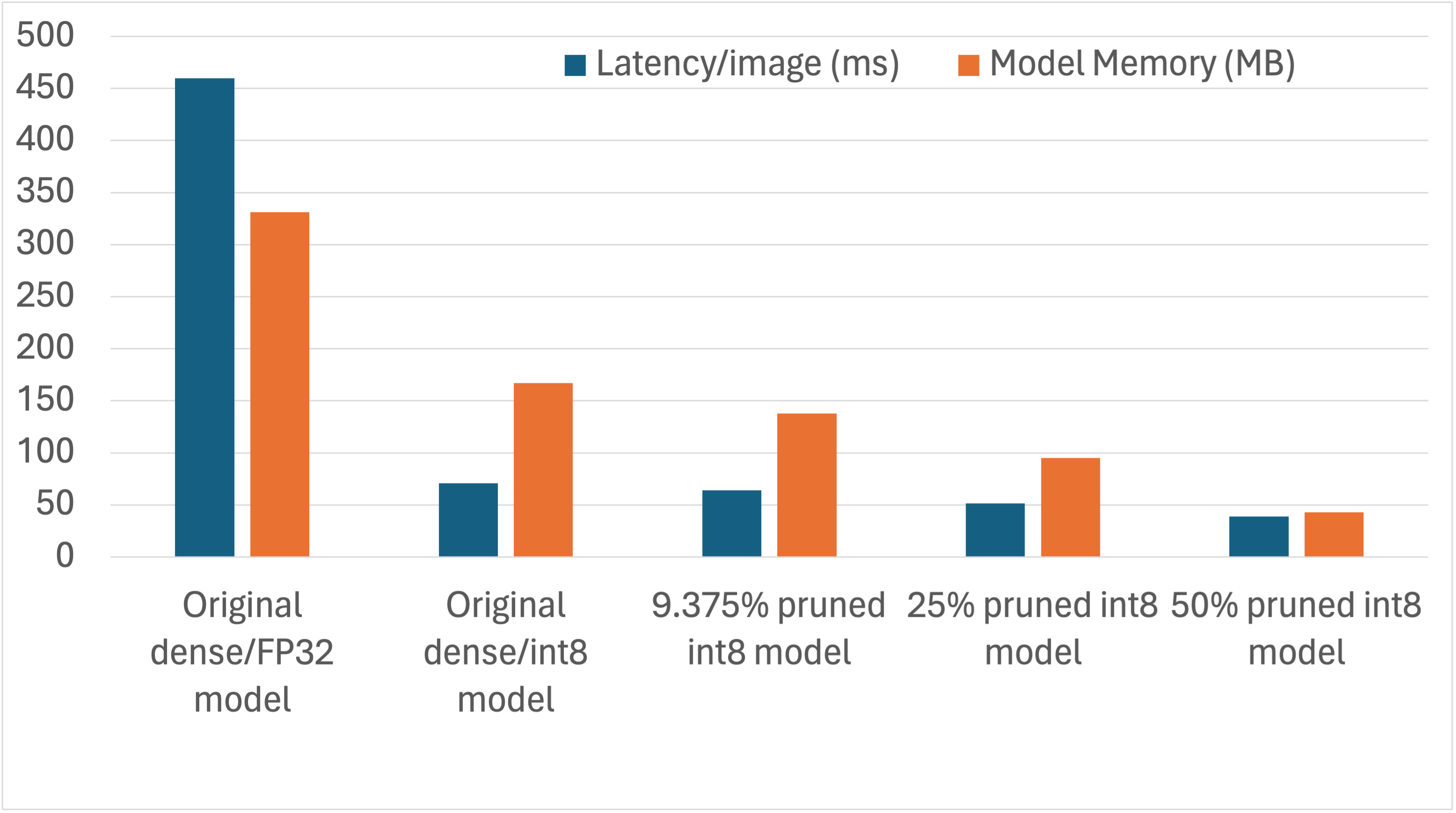}\vspace{-0.2in}
    \caption{Latency and memory results for deit\_base\_patch16 model with varying pruning and quantization. All models are scripted, mobile optimized, and converted to Pytorch Lite format.\vspace{-0.2in}}
    \label{fig:base_results}
\end{figure}

\begin{figure*}[!t]
    \centering
    \includegraphics[width=1\textwidth]{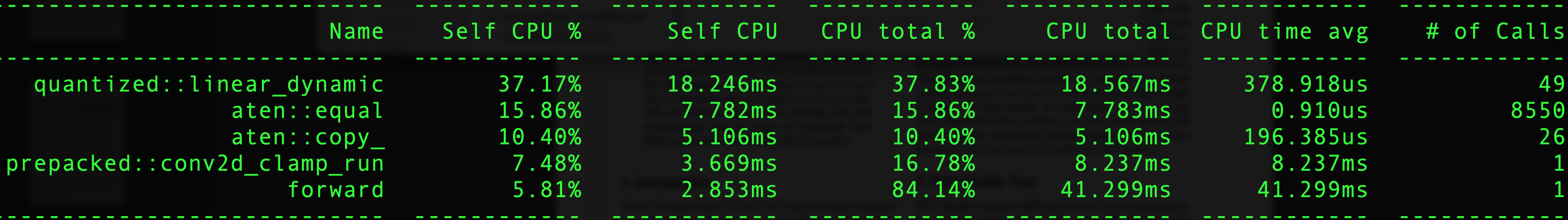}\vspace{-0.2in}
    \caption{Latency profile for deit3\_medium\_patch16 int8 model at 9.375\% pruning level for x86 backend. The model is scripted, mobile optimized, and converted to Pytorch Lite format.\vspace{-0.2in}}
    \label{fig:profile}
\end{figure*}

\begin{figure}[t]
    \centering
    \includegraphics[width=1\columnwidth,trim=4 4 4 4,clip]{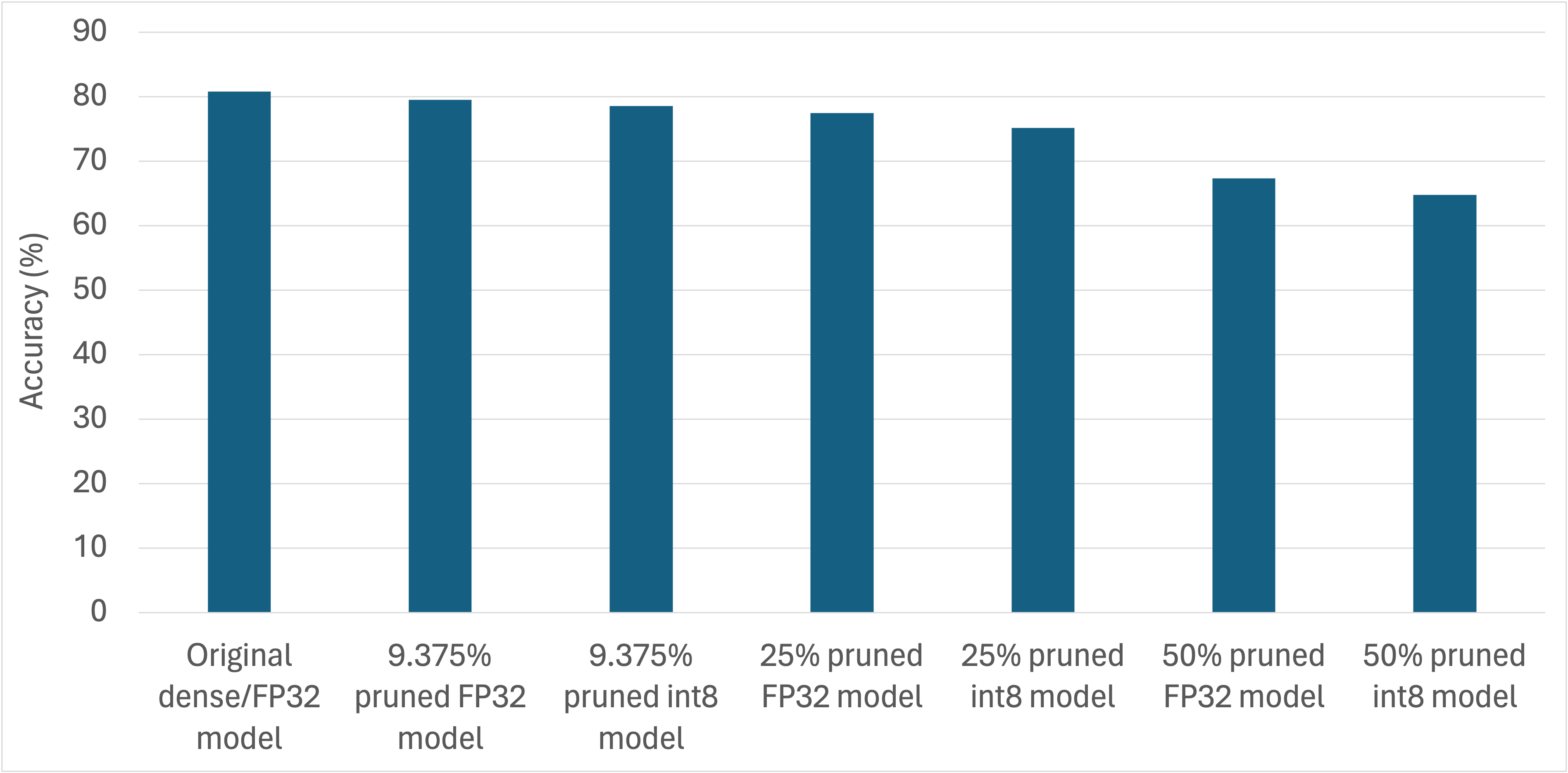}\vspace{-0.2in}
    \caption{Accuracy results for deit\_base\_patch16 model with varying pruning and quantization. All models are scripted, mobile optimized, and converted to Pytorch Lite format.\vspace{-0.2in}}
    \label{fig:base_acc}
\end{figure}

\begin{figure}[t]
    \centering
    \includegraphics[width=1\columnwidth,trim=4 4 4 4,clip]{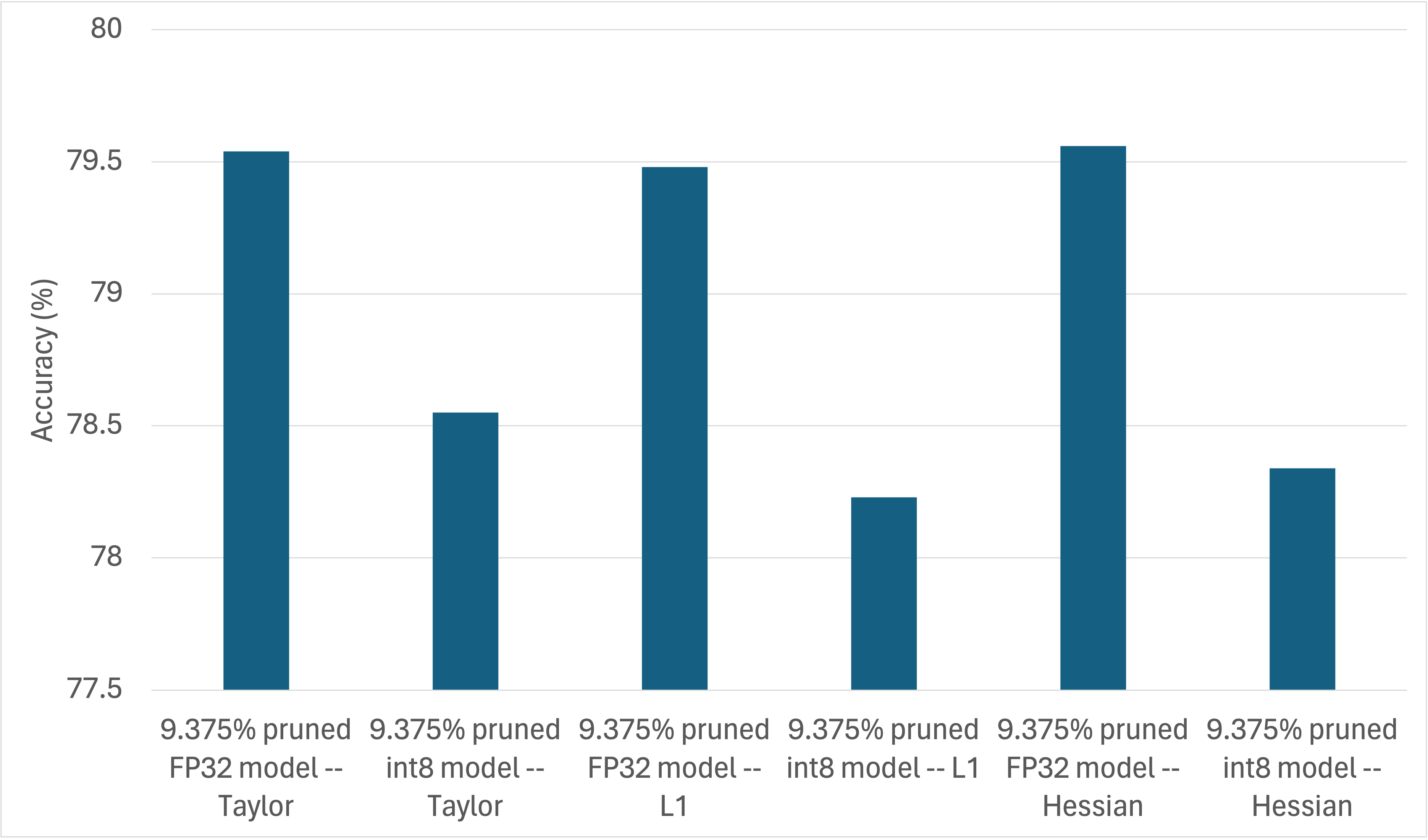}
    \caption{Accuracy results for deit\_base\_patch16 model with different structured pruning groupings (for both FP32 and int8 models). All models are scripted, mobile optimized, and converted to Pytorch Lite format.}
    \label{fig:pruning}
\end{figure}

In this section, we demonstrate the effectiveness of PQV-Mobile to prune and quantize Facebook's Data Efficient Image Transformers (DeiT). We evaluate the models' latency-memory-accuracy trade-offs. The structured pruning importance used in this study is based on Taylor's expansion (as shown later, it performs the best in terms of accuracy). The pruned model is then finetuned for 60 epochs (using distributed training on 4 GPUs) at a learning rate of 0.00015 with a batch size of 64. The finetuned pruned models are then quantized for the x86 backend engine and then converted to optimized and Lite format. We use the ImageNet dataset~\cite{imagenet} for image classification tasks and run the models on Intel Xeon ES2695 at 2.3 GHz. Pytorch-2.0.0 is used in all our experiments. Please note that our experiments are only meant to demonstrate the capabilities of our tool and not to achieve the state-of-the-art accuracy. All comparisons are performed on scripted, mobile optimized, and Pytorch Lite format models.

Figure~\ref{fig:base_results} shows latency/image and memory for deit\_base\_patch16 model after varying degrees of pruning and quantizing the model from FP32 to int8. Quantizing the original dense model to int8 leads to $6.47\times$ lower latency as the Pytorch Lite interpreter is more effective with quantized models than FP32 models. Pruning the quantized model by 9.375\% leads to further 9.8\% lower latency (an overall $7.14\times$ latency reduction over the original dense/FP32 Lite model). Increasing the amount of pruning to 25\% and 50\% shows further improvements in latency by 27.9\% and 45.4\%, respectively. The tool can also be used to perform a detailed profiling of the latency of the mobile optimized and Lite model as shown in Figure~\ref{fig:profile}, where we can identify the bottleneck based on the time spent on the various operations (as depicted by the Name of the process). In terms of accuracy, we found it to be similar across the different backends.

While pruning the model to 50\% leads to significant improvements in latency and memory, Figure~\ref{fig:base_acc} shows that the accuracy degradation is considerable. Pruning 
the original model by 9.375\% shows 1.25\% lower accuracy. Further quantizing this model leads to an additional 0.99\% loss in accuracy. While 25\% pruned int8 model achieved a speedup of 27.9\% over 9.375\% pruned int8 model, its accuracy loss is 3.41\%. These results demonstrate the importance of performing latency-memory-accuracy trade-offs which can be seamlessly performed using our PQV-Mobile tool. 

Figure~\ref{fig:pruning} shows how the different types of structured pruning groupings affect accuracy and motivates why we chose Taylor pruning for all of our experiments. We prune the dense model by 9.375\%, finetune it, and also quantize the finetuned pruned model to int8 for this experiment. Although there is a very small change in accuracy between Taylor, L1-norm, and Hessian-based pruning, Taylor outperforms the other methods.

\begin{figure}[!t]
    \centering
    \includegraphics[width=1\columnwidth,trim=4 4 4 4,clip]{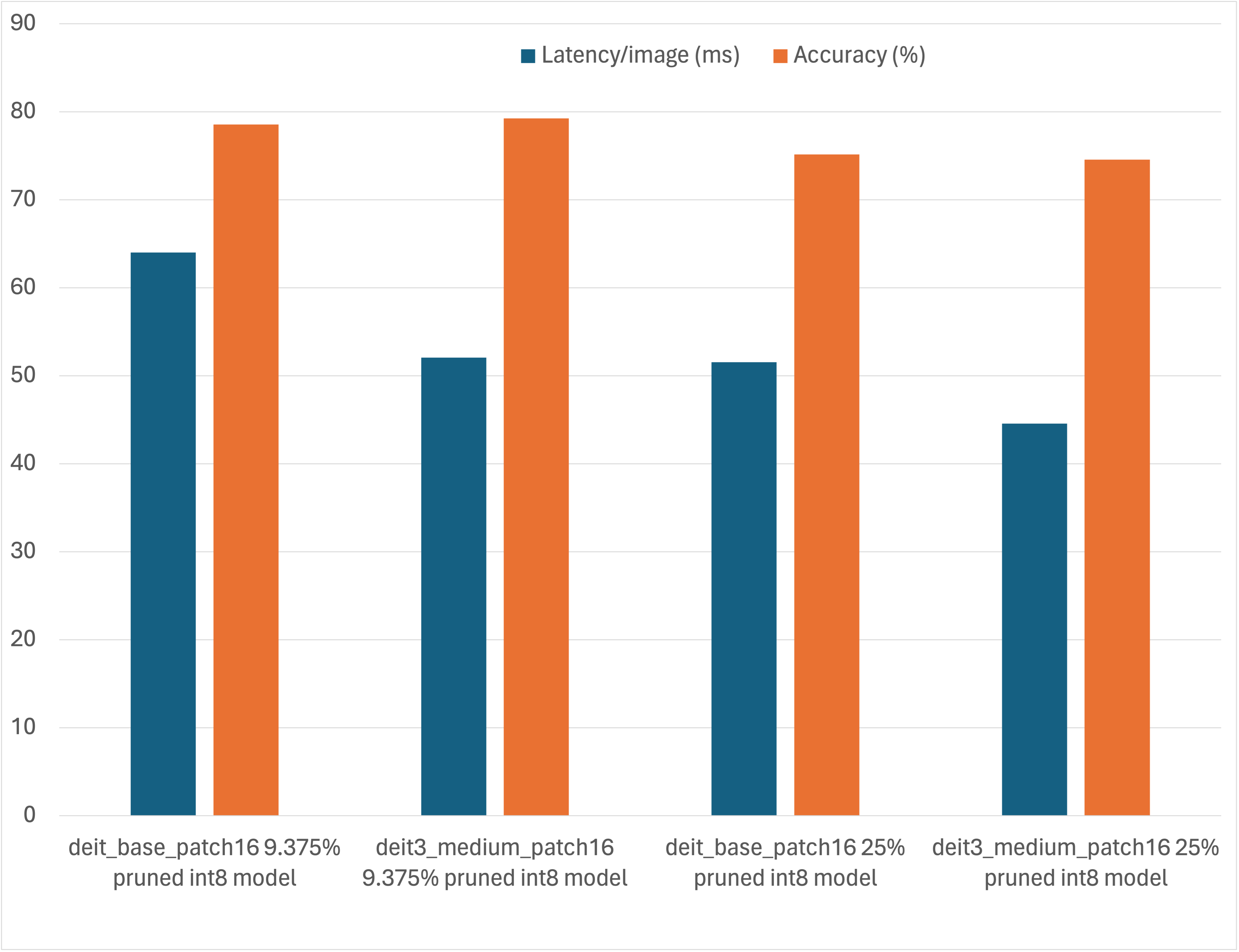}
    \caption{Latency and accuracy results for deit\_base\_patch16 model vs. deit3\_medium\_patch16 model with varying pruning levels. All models are scripted, mobile optimized, and converted to Pytorch Lite format.}
    \label{fig:models_comp}
\end{figure}

We further compare the latency and accuracy of pruning and quantizing deit\_base\_patch16 model with deit3\_medium\_patch16 model (Figure~\ref{fig:models_comp}). The original dense FP32 accuracies for the two models are 80.79\% and 82.19\%, respectively. The latter is a smaller model with 38.85M parameters compared to 86.56M parameters in the former. As evident, the models show similar accuracy when pruned to the same levels (9.375\% or 25\%) at int8 quantization. However, deit3\_medium\_patch16 model shows latency improvements by 18.65\% and 13.55\%, respectively over deit\_base\_patch16 model.

\begin{figure}[!t]
    \centering
    \includegraphics[width=0.7\columnwidth,trim=4 4 4 4,clip]{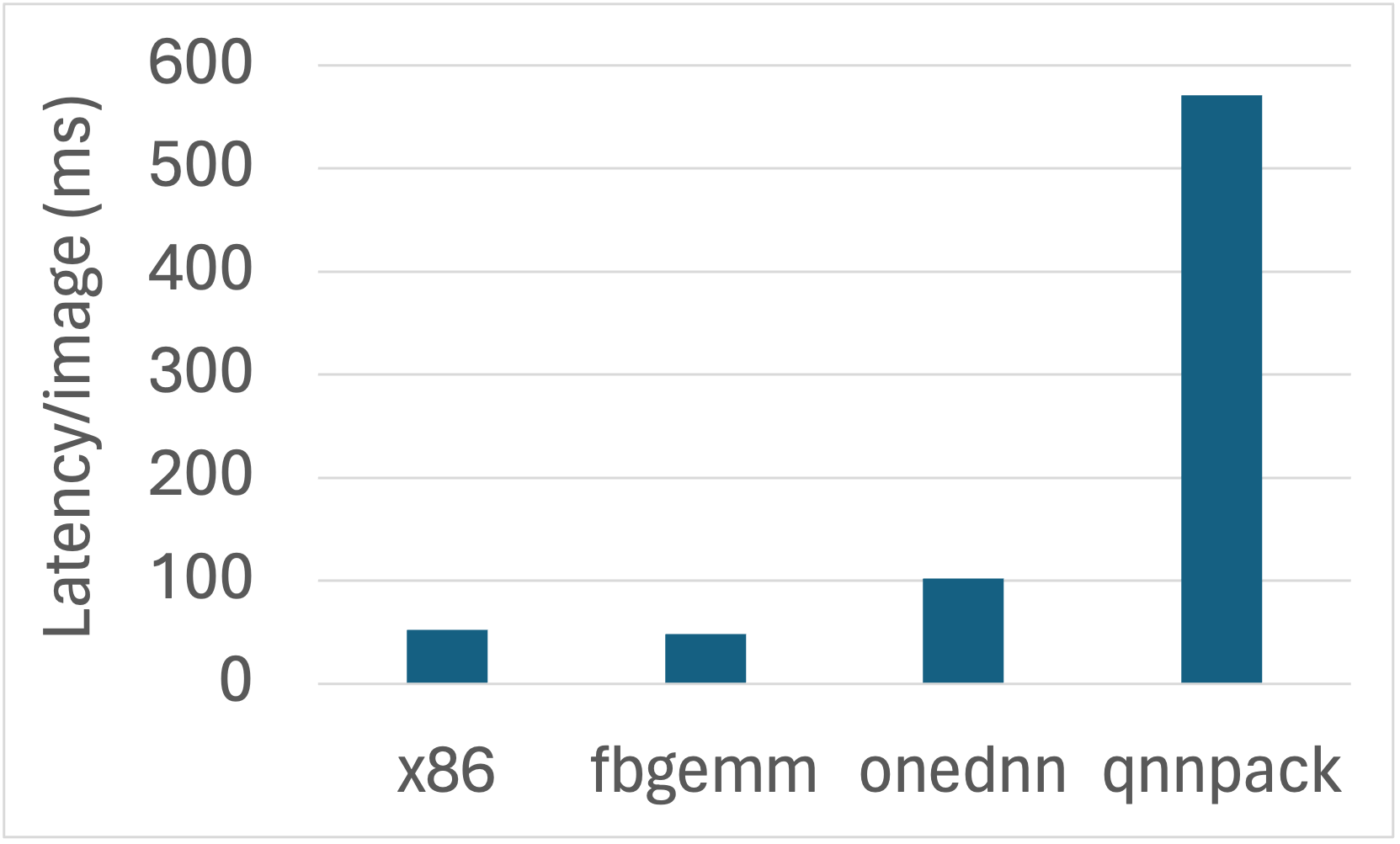}
    \caption{Latency results for deit3\_medium\_patch16 int8 model at 9.375\% pruning level for different hardware backends. All models are scripted, mobile optimized, and converted to Pytorch Lite format.}
    \label{fig:backends}
\end{figure}

Finally, as shown in Figure~\ref{fig:backends}, we also evaluate the latency for different int8 quantization hardware backends. We use the 9.375\% pruned deit3\_medium\_patch16 model for this experiment. We find that x86 and FBGEMM backends to be the best with FBGEMM slightly outperforming x86. 
These results are expected as we are running on an x86 machine with Advanced Vector Extensions (AVX) enabled, which are used for fast path executions for both x86 and FBGEMM backends.

\section{Conclusion and Future Work}
\label{sec:conclusion}

This paper presents a combined pruning and quantization tool, called PQV-Mobile, to optimize vision transformers for mobile applications. The tool is able to support different types of structured pruning based on magnitude importance, Taylor importance, and Hessian importance. It also supports quantization from FP32 to FP16 and int8, targeting different mobile hardware backends. We demonstrate the capabilities of our tool and show important latency-memory-accuracy trade-offs for different amounts of pruning and int8 quantization with two types of Facebook DeiT models.

As future work, we plan to extend PQV-Mobile to int4 quantization. Additionally, we will extend this tool to target large language models as well.

\section{Acknowledgements}

This work was performed under the auspices of the U.S. Department of Energy by LLNL under contract DE-AC52-07NA27344 (LLNL-CONF-865054).


\bibliography{main}
\bibliographystyle{icml2023}

%

\end{document}